\documentclass[runningheads]{llncs}

 \usepackage{eccv}

\usepackage{eccvabbrv}

\usepackage{graphicx}
\usepackage{booktabs}

\usepackage[accsupp]{axessibility}  

\usepackage[breaklinks,colorlinks,citecolor=eccvblue]{hyperref}

\usepackage{orcidlink}

\newcommand{\name}{EmoCAM}
\newcommand{\oi}{Open Images}

\begin{document}

\title{\name{}: Toward Understanding What Drives CNN-based Emotion Recognition} 

\titlerunning{\name{}}

\author{Youssef Doulfoukar*\inst{1} \and
Laurent Mertens*\inst{1,2}\orcidlink{0000-0001-5175-2673} \and
Joost Vennekens\inst{1,2,3}\orcidlink{0000-0001-5175-2673}}

\authorrunning{Y.~Doulfoukar \etal}

\institute{KU Leuven, De Nayer Campus, Dept. of Computer Science \\
  J.-P. De Nayerlaan 5, 2860 Sint-Katelijne-Waver, Belgium \and
  Leuven.AI - KU Leuven Institute for AI, 3000 Leuven, Belgium \and
Flanders Make@KU Leuven, 3000 Leuven, Belgium \\
\email{laurent.mertens@kuleuven.be} \\
\small{*Authors contributed equally}
}

\maketitle

\begin{abstract}
Convolutional Neural Networks are particularly suited for image analysis tasks, such as Image Classification, Object Recognition or Image Segmentation. Like all Artificial Neural Networks, however, they are ``black box'' models, and suffer from poor explainability. This work is concerned with the specific downstream task of Emotion Recognition from images, and proposes a framework that combines CAM-based techniques with Object Detection on a corpus level to better understand on which image cues a particular model, in our case EmoNet, relies to assign a specific emotion to an image. We demonstrate that the model mostly focuses on human characteristics, but also explore the pronounced effect of specific image modifications.
\end{abstract}

\section{Introduction}
\label{s:intrso}
Thanks to recent progress, image analysis problems such as Object Detection using Artifical Neural Networks (ANN) can be more or less considered to be solved \cite{8627998, KAUR2023103812}. However, higher-order tasks, such as identifying the emotion content of an entire image, remain more challenging. Convolutional Neural Network (CNN) models such as EmoNet \cite{EmoNet19} present a promising approach in this area, but its results are not yet completely convincing. This raises the question to which extent this network is actually picking up meaningful cues in the images, and to what extent it is learning spurious correlations that may be present in the private dataset on which it was trained.

ANNs are still considered ``black box'' models, and the domain that attempts to untangle how they make the predictions they make, i.e., to improve their \emph{explainability}, is a very active one \cite{SAEED2023110273, Yang2023SurveyOE, ALI2023101805}. One of the techniques for this is Class Activation Maps \cite{CAM}, or CAM, which allows to highlight those parts of the image that contributed most to a model's (say, a CNN image classifier) output. This technique allows to visually inspect individual images or videos, but does not immediately allow for an automated global analysis on a corpus level.

To answer the earlier question of what image cues a CNN-based Emotion Recognition network, in casu EmoNet, most relies on, we propose \name{}. Our framework combines two information streams, namely CAM and Object Detection, to build a pipeline that allows to determine those object classes that most contributed to the model's decision making on a corpus level. Besides better understanding what object classes the model relies most upon, we also want to explore the potential of applying minor changes to the input images that steer the model towards a specific emotion by leveraging the obtained information from our \name{} analysis. Our source code can be found at \url{https://gitlab.com/EAVISE/lme/emocam}.

The remainder of this paper is organized as follows: in \cref{s:methodology} we describe our proposed framework in detail, followed by \cref{s:results} where we look at a concrete case using the EmoNet network and FindingEmo \cite{FindingEmo24} image dataset; limitations and roads for future work are explored in \cref{s:limitations} and we present concluding remarks in \cref{s:conclusion}.

\section{Methodology}
\label{s:methodology}
\begin{figure}[tb]
  \centering
  \includegraphics[width=10cm]{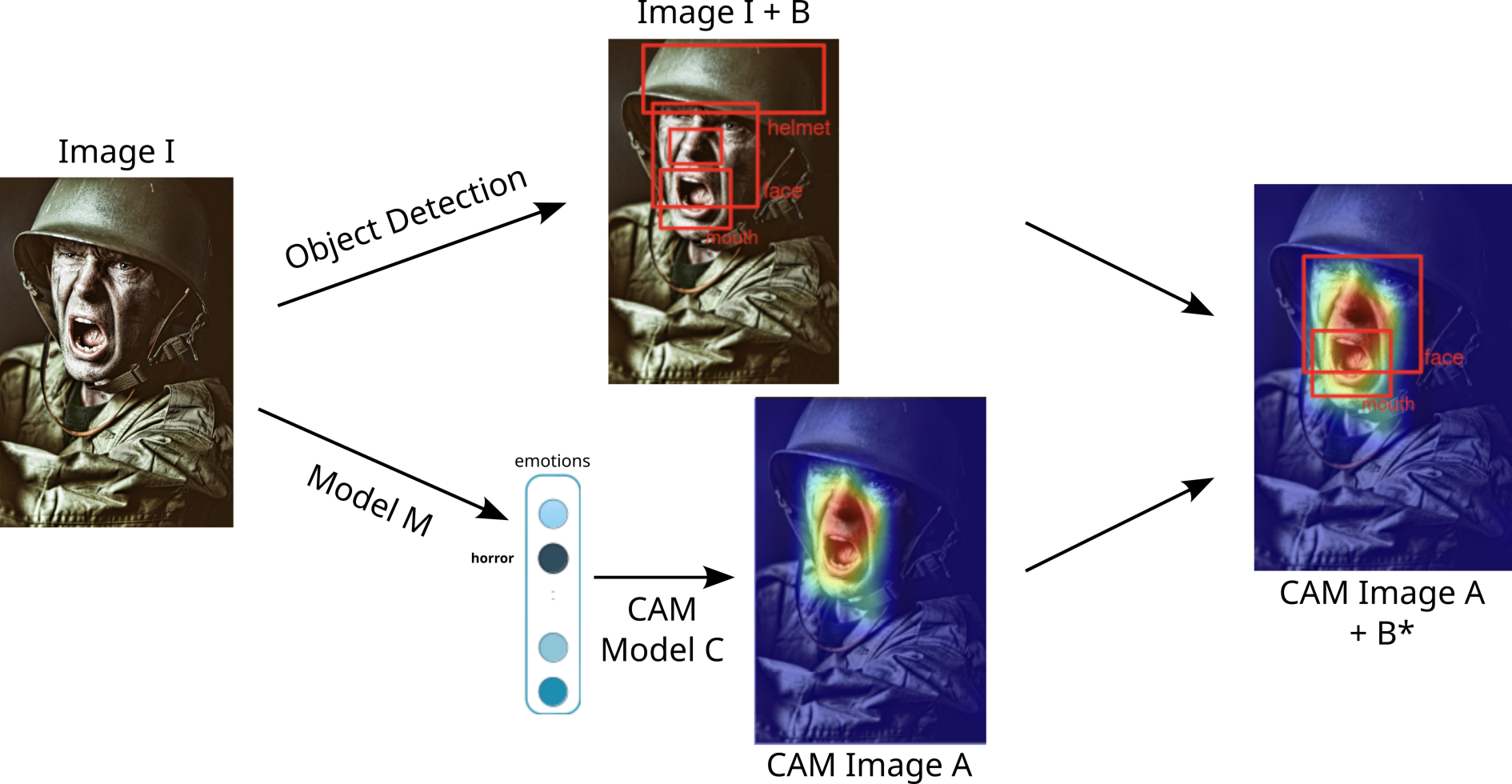}
  \caption{Schematic illustrating pipeline combining CAM with Object Detection. Photo by Sander Sammy on Unsplash, \url{https://tinyurl.com/2tc69hfy}, \href{https://unsplash.com/license}{Unsplash} license.}
  \label{fig:pipeline_single_figure}
\end{figure}

We start our analysis by applying, for a given CNN model $M$ and corpus $D$, the following steps to each image $I \in D$, schematically illustrated in \cref{fig:pipeline_single_figure}.

First, we process $I$ with the object detection network of our choice, in casu, YOLOv3 \cite{YOLOv3} trained on the \oi{} dataset.\footnote{We use the PyTorch LightNet \cite{lightnet18} implementation, and OpenImages weights available from \url{https://pjreddie.com/darknet/yolo/}.} We opted for this particular pretrained network as other popular Object Detection dataset choices such as PASCAL VOC (20 classes) and MS-COCO (80 classes) are too restricted in the classes they propose. By contrast, \oi{}, which contains 601 classes, presents a nice balance between human-related classes (e.g., ``human face'', ``mouth'', etc.), and more general classes representing contextual elements (e.g., ``car'', ``tree'', etc.). The result of this operation is a list of detected objects and their corresponding bounding boxes $B$. We filter the YOLOv3 output by keeping only bounding boxes with an IoU score $>0.005$.

Second, we process $I$ with $M$, and apply a CAM-based technique $C$ to the last convolutional layer of $M$.\footnote{We experimented with other layers, as well as combination of layers, but the best results were obtained using only the last layer.} This gives us an activation map that we overlay on top of $I$ to obtain a new image $A$.

Finally, we lay the bounding boxes $B$ on top of $A$, and look for those boxes $b \in B$ for which the average CAM activation, or \emph{importance}, $C_{Act} > 0.3$, with $C_{Act}$ defined as the sum of the CAM activations within the box divided by the area of the box.\footnote{To be clear, for this we take the activation values from the original grayscale CAM output, not from the colorized version of the output.} The threshold was heuristically determined by visually inspecting a limited set of images. We refer to these boxes as the set $B^*$, and interpret these as those objects that most contributed to the model's decision.

Once we have found $B^*$ for every image in our corpus, we then analyse these data to find associations between object classes and output labels by constructing an association matrix $M_A$, where $a_{ij} \in M_A$ represents the number of images labeled with the $j^{th}$ EmoNet emotion label in which the $i^{th}$ object class has been detected at least once. By dividing each column $j$ through by the total number of images labeled with the $j^{th}$ emotion such as to obtain percentages (after doing $\times 100$), we obtain $M^\prime_A$ which allows to ignore imbalances in the prediction rates of the different emotions.

\section{Results}
\label{s:results}
We tested our proposed approach using the EmoNet model and the FindingEmo dataset. EmoNet is a model obtained through replacing the last layer of an AlexNet model pretrained on the ImageNet \cite{deng2009imagenet} corpus. This last layer was then trained on a private dataset of 137,482 images annotated for the emotion they evoke in the observer with one of 26 custom emotion labels. We use the Python port by L. Mertens \cite{EmoNet-Py} of the original Matlab release.

FindingEmo is an image dataset consisting of 25,869 images annotated for, a.o., the dominant emotion in the picture, using one of the 24 emotion labels in Plutchik's Wheel of Emotions \cite{PLUTCHIK19803}. All images represent multiple people in various natural settings and with varying degrees of interaction among them.

We first present detailed results for Grad-CAM \cite{GradCAM} in \cref{ss:gradcam}, and follow this up with an exploration of the effect of using other CAM-based methods in \cref{ss:cam_rsa}.\footnote{For the CAM analysis, We use the Python packages \texttt{grad-cam} \cite{jacobgilpytorchcam} and \texttt{captum} \url{https://pypi.org/project/captum/}.} Finally, we briefly explore the effect on the predicted label of artificially adding certain objects to images, attempting to answer the question whether the presence of certain objects can \emph{cause} a specific label to be predicted.

\subsection{Results for Grad-CAM}
\label{ss:gradcam}
A heatmap depicting $M^{\prime}_A$ as obtained using Grad-CAM together with EmoNet applied on the FindingEmo corpus can be found in \cref{fig:cam_rsa}. We limit ourselves to the 25 most prominent \oi{} classes (as determined by the average of the corresponding row in $M^{\prime}_A$). A clear conclusion to be drawn from this graph is that human features do indeed contribute the most to the decision making, most particularly the human face which, except for ``Clothing'', represents the most important class for each EmoNet label.

\begin{figure}[tb]
  \centering
  \includegraphics[width=\linewidth]{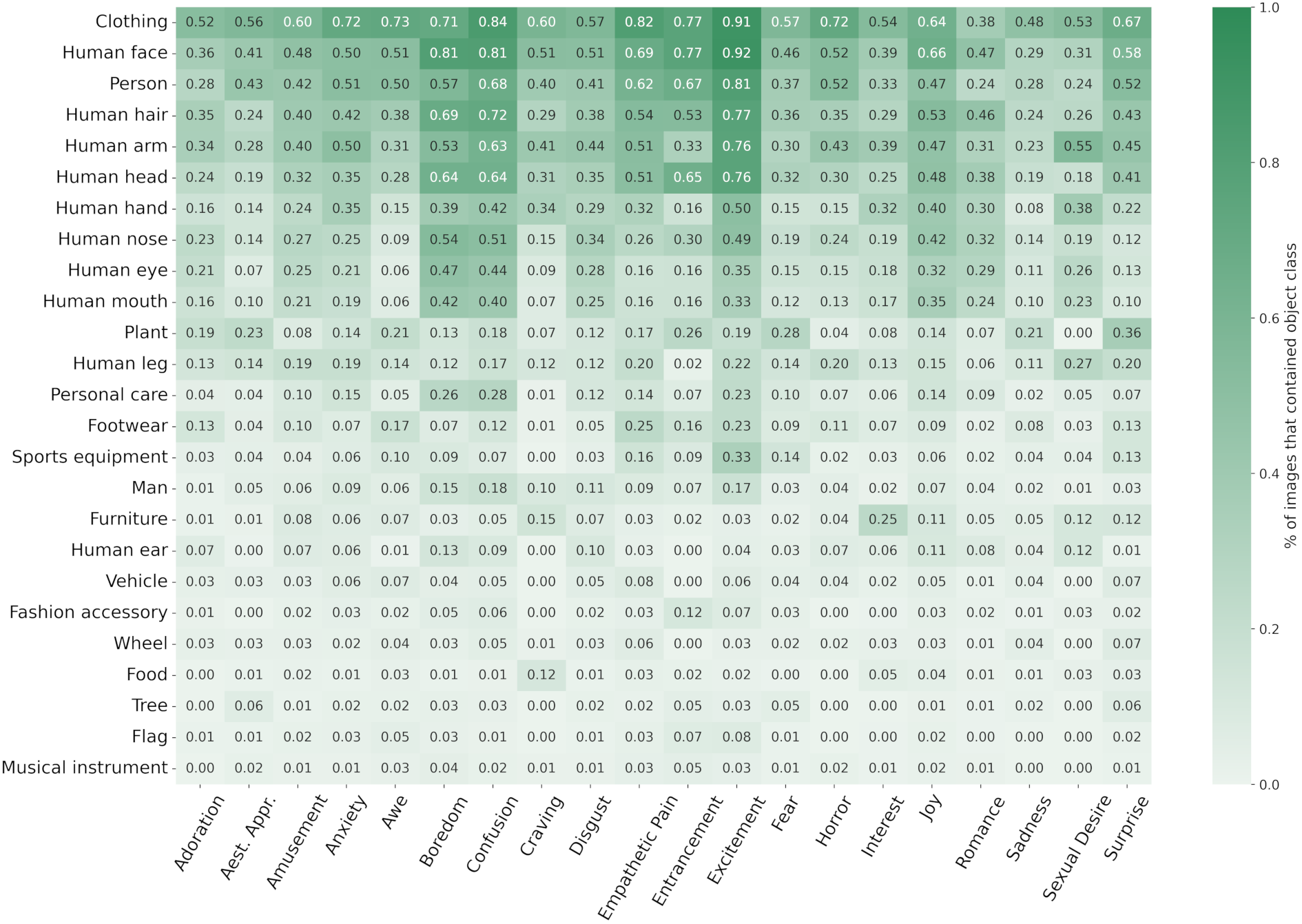}
  \caption{Association between \oi{} classes and predicted EmoNet label. Heatmap entries represent the percentage of images labeled with a certain EmoNet label for which at least one object of the corresponding \oi{} class was detected with high enough importance. ``Aest. Appr.'' = Aesthetic Appreciation.}
  \label{fig:cam_rsa}
\end{figure}

Additionally, some more specific associations do manifest themselves. Clear examples are the association between ``Sports equipment'' and ``Excitement'', and ``Food'' and ``Craving'', both of which seem logical. Less clear is, e.g., the association between ``Furniture'' and ``Interest'', or ``Plant'' and ``Surprise'', which hint of spurious associations resulting from biases in either or both the EmoNet training dataset and FindingEmo. 

\subsection{Comparison of CAM Methods}
\label{ss:cam_rsa}
To answer the question to what extent different CAM methods yield different results, we performed a Representational Similarity Analysis \cite{RSA} as follows. For each CAM method $C \in \{$Grad-CAM, Ablation-CAM\cite{AblationCAM}, LIME\cite{Lime}, LRP\cite{LRP}, LIFT-CAM\cite{Lift}$\}$, we determine the association matrix $M_A$ with all \oi{} classes as described in \cref{s:methodology}, keeping the same emotion and class ordering for each $C$\footnote{Which ordering is used does not matter, as long as it is consistently used.}. We then flatten each matrix by concatenating all rows, turning it in to a 1D vector $V_{M_C}$. Finally, we construct a matrix $R$ where each entry $R_{C{C^\prime}}$ represents the Spearman Correlation rank between $V_{M_C}$ and $V_{M_{C^\prime}}$. The resulting matrix is shown in \cref{fig:cam_rsa}. All related $p$-values were $<\!\!< 0.05$, indicating statistical significance.

\begin{figure}[tb]
  \centering
  \includegraphics[width=6cm]{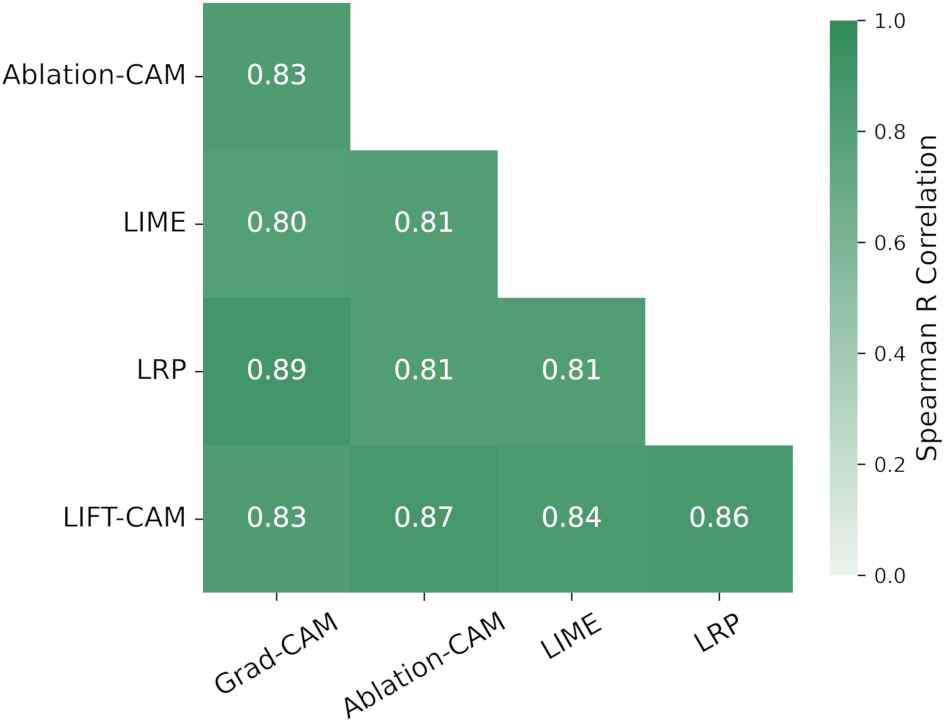}
  \caption{RSA analysis of different CAM methods.}
  \label{fig:cam_rsa}
\end{figure}

The consistently high correlation values between all pairs indicate that variations in results obtained through different CAM methods can be expected to be minimal. We did observe both LIFT-CAM and LRP resulting in a notable association between ``Pillow'' and ``Sexual Desire''. Other than this, the differences between the methods appear to lie within the relative strengths of the associations observed, rather than the assocations themselves.

\subsection{Prediction Stability}
\label{ss:adv_ex}
\begin{figure}[tb]
  \centering
  \includegraphics[width=8cm]{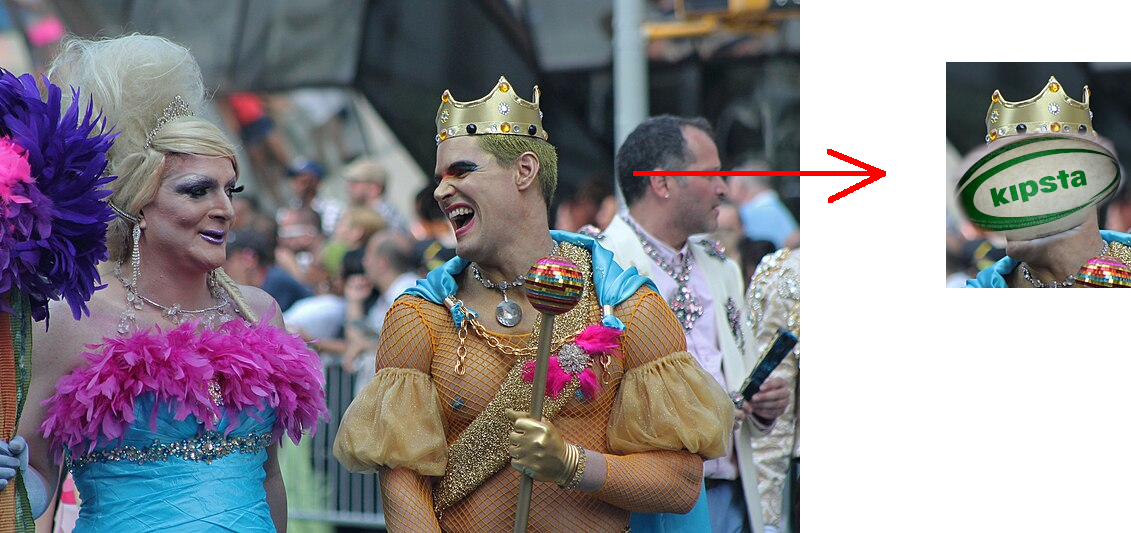}
  \caption{Adversarial example. Original image on the left, labeled by EmoNet as 92.9\% ``Joy''. Modified part on the right; full modified image labeled as 66.1\% ``Excitement''. Sources: original photo by istolethetv,  \url{https://tinyurl.com/3r86e3w2}, CC 2.0 license; Rugby ball by Peter Griffin, \url{https://tinyurl.com/dmb77rks}, CC0 license.}
  \label{fig:adv_ex}
\end{figure}

To illustrate how the obtained knowledge can be applied to fool the network by creating an adversarial attack, consider the image shown in \cref{fig:adv_ex}. We know from \cref{ss:gradcam} that there is a high association between the object category ``Sports equipment'' and EmoNet label ``Excitement''. This inspired us to take an image labeled with high probabilty as ``Joy'' (92.9\%; Excitement: 1.8\%). After altering this image by pasting a rugby ball on top of the head of one of the two main subjects, the prediction changes to 66.1\% ``Excitement'' (``Joy'': 30.2\%), demonstrating the dramatic effect the presence of a particular object can have on the model's output.

Note that the position of the pasted object greatly influences the effect it has. Moving the rugby ball to the immediate right of the subject's face alters the predictions to 43.9\% ``Joy'' and 42.1\% ``Excitement'', while moving it to the immediate left only alters the predictions by 4\% in the same directions (``Joy'': 34.2\%; ``Excitement'': 62.1\%). Covering the other subject's head instead, we obtain 52.0\% ``Joy'' and 30.7\% Excitement.

\begin{figure}[tb]
  \centering
  \includegraphics[width=11cm]{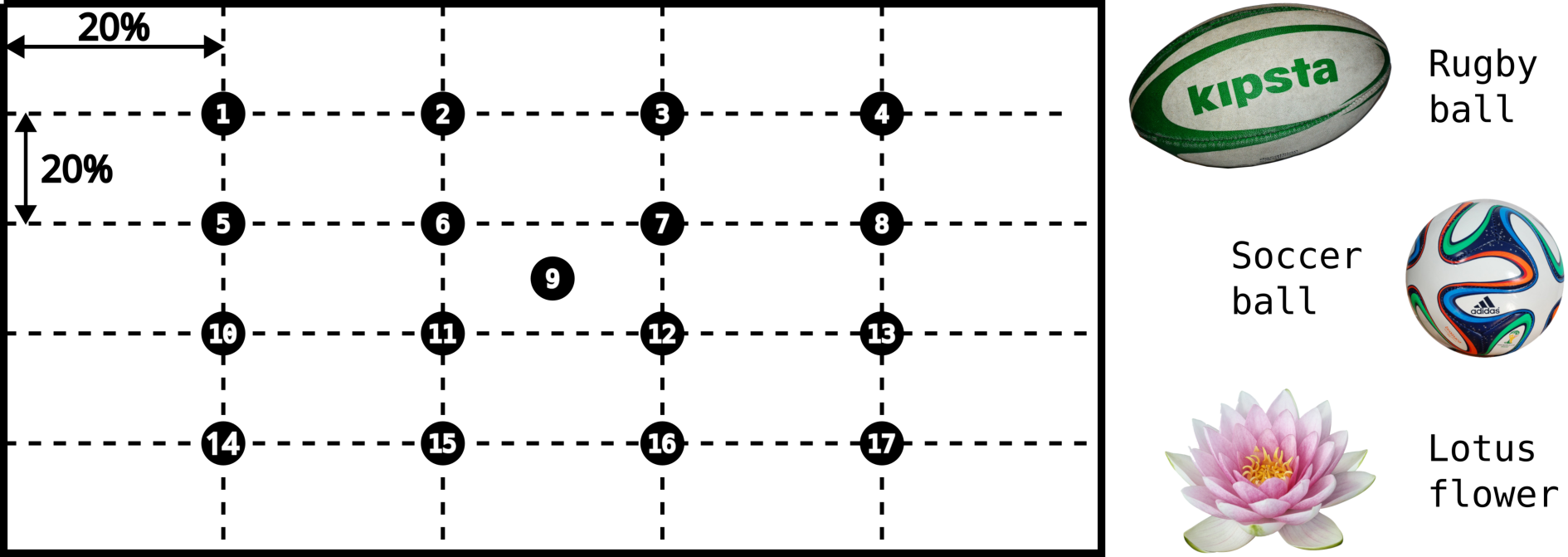}
  \caption{Schematic illustration of ``paste object in image'' experiment. The grid on the left illustrates the relative positions within the image the objects are pasted and centered at, with the considered objects shown on the right. Sources: Rugby ball, see \cref{fig:adv_ex}; Soccer ball by Jean Schecter, \url{https://tinyurl.com/3xxkkysb}, CC 4.0 BY-NC; Lotus flower at \url{https://pngimg.com/image/69752}, CC 4.0 BY-NC.}
  \label{fig:schema_exp}
\end{figure}

\begin{figure}[tb]
  \centering
  \includegraphics[width=\linewidth]{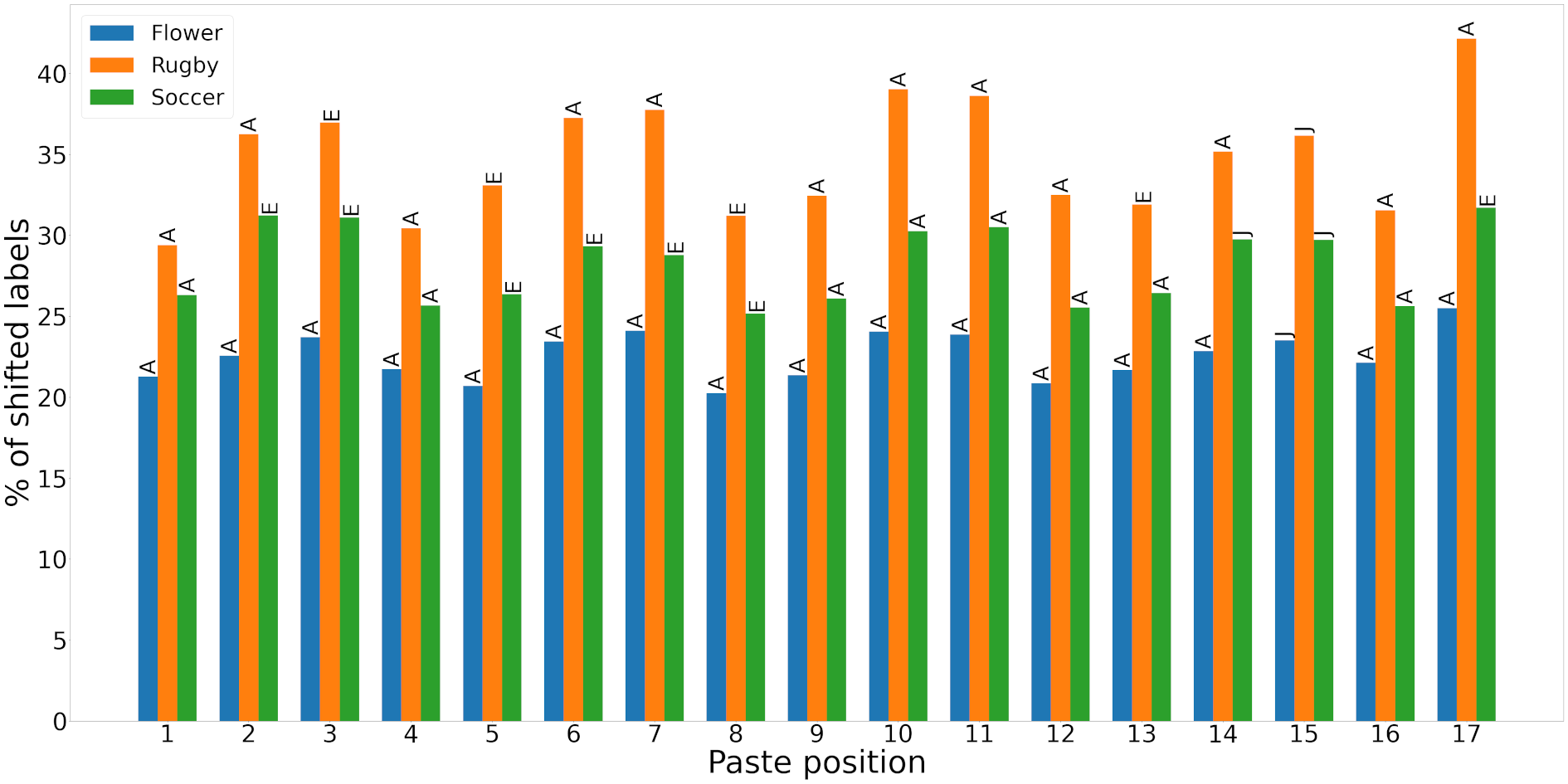}
  \caption{Percentage of images in the corpus whose predicted label changed when a specific object was pasted at a specific position. Refer to \cref{fig:schema_exp} for the positions and objects. The letters `A' for `Admiration', `E' for `Excitement' and `J' for `Joy' atop each bar indicate the EmoNet label most often switched to.}
  \label{fig:barchart_exp}
\end{figure}

We further investigate this effect by performing the following experiment. For each $I$ in $D$, we paste a given object $O \in \{$Rugby ball, Soccer ball, Lotus flower$\}$, resized such that its height equals $0.2~\times~\mbox{height}(I)$, in $I$ centered at each one of a set of predefined relative positions $P$ within the image, resulting in $size(P)$ alterations to $I$. The positions and objects considered are illustred in \cref{fig:schema_exp}. We then send the altered images through EmoNet, and observe how the prediction was affected. The results are shown in \cref{fig:barchart_exp}.

The results confirm that the model shows high sensitivity to certain objects. Although differences of up to more than 10\% can be observed in \cref{fig:barchart_exp}, specifically for the rugby ball, no real tendencies reveal themselves. In combination with the example in \cref{fig:adv_ex}, we hypothesize the differences are not so much due to the \emph{absolute} position of the object, but to what it \emph{occludes}. For the rugby ball, 4 out of 17 positions resulted most often in a label switch to ``Excitement''. For the soccer ball, the number increases to 7. The Lotus flower clearly results in much less label shifts overall, with not a single position favoring ``Excitement'', confirming the importance of the object class in effecting a label switch.

\section{Limitations and Future Work}
\label{s:limitations}
Although the currently described approach already provides valuable insights, some limitations are to be noted.

First, the approach is, by definition, heavily dependent on the choice of Object Detection network and its corresponding classes and performance. The upside is that, as a plug-and-play component, different Object Detection networks can be chosen for different tasks, allowing to pick object classes tailored to the task at hand.

\begin{figure}[tb]
  \centering
  \includegraphics[width=5cm]{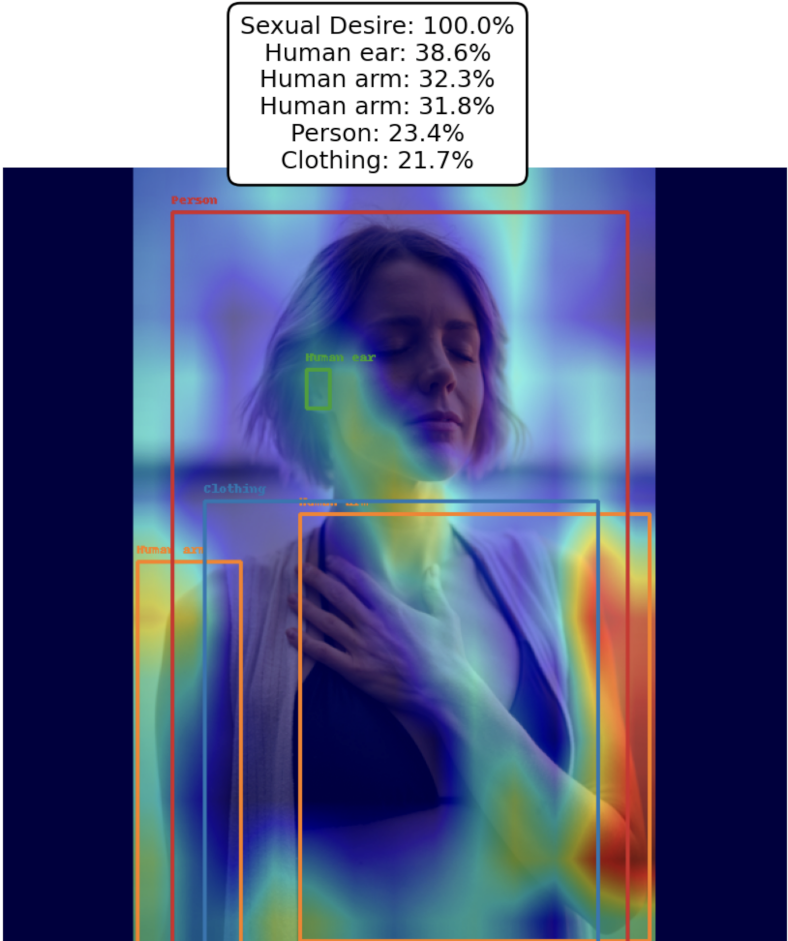}
  \caption{An instance were the current \name{} approach fails to detect the most important class. Source: original image by Darius Bashar on Unsplash, \url{https://tinyurl.com/4eephaxb}, \href{https://unsplash.com/license}{Unsplash} license.}
  \label{fig:limitations_1}
\end{figure}

Second, our current implementation does not take into account the size of the bounding boxes, which can result in suboptimal results. Consider, e.g., the example shown in \cref{fig:limitations_1}. Although the subject's ear is clearly not the most important contributing element in the picture, because of the small size of the ``Human ear'' bounding box the average CAM activation is nonetheless the highest, spuriously pushing this object class to the top. Two main paths could be explored to counter this issue. The most straightforward would be to develop a scoring function that does take into account the bounding box size, or the activation distribution within it. Alternatively, segmentation models could be used instead of bounding box detection models, so as to obtain clearly delineated zones representing the different objects. Then of course, our first limitation still applies, i.e., the segmentation classes need to be relevant to the task at hand.

Third, with regard to our experiment described in \cref{ss:adv_ex}, a more interesting approach might be to, instead of, or along with, considering only a fixed set of positions, paste the object at the center of bounding boxes relating to specific features such as human heads, thus investigating the effect of masking specific objects. We also intend to apply EmoCAM to the modified images to explore if the shift in label is reflected in a shift in focus in the modified image.

\section{Conclusion}
\label{s:conclusion}
We propose the novel \name{} approach to explaining CNN decisions, specifically with the downstream task of Emotion Recognition from images in mind. Our objective is threefold: 1) better understanding what parts of the input image the model uses to make its decision, 2) allowing to check whether or not the information used by the model aligns with expectations from a human perspective, and 3) uncovering potential model biases. We have demonstrated our approach using the EmoNet model, FindingEmo dataset and multiple CAM techniques. Using our approach, we found that EmoNet indeed shows a strong focus on human elements, most notably (parts of) the human face, which is encouraging as it aligns with our understanding of human emotion recognition from Psychology. Nevertheless, we also found the model output to be quite unstable, in that adding specific objects (e.g., a rugby ball) to an image can dramatically alter its output and steer it towards a specific target emotion (e.g., ``Excitement'').

%
%
\bibliographystyle{splncs04}
\bibliography{EmoCAM}

\end{document}